\documentclass[letterpaper, 10pt, conference]{ieeeconf}  

\IEEEoverridecommandlockouts                              

\usepackage[dvipsnames]{xcolor}
\usepackage{graphicx}
\usepackage{subcaption}
\usepackage{amsmath}
\usepackage{amssymb}
\usepackage{wrapfig}
\usepackage{paralist}
\usepackage{multirow}
\usepackage{float}
\usepackage{booktabs}
\usepackage{makecell}
\usepackage{url}
\usepackage[normalem]{ulem}
\usepackage{hyperref}

\usepackage{tikz}
\usetikzlibrary{positioning,shapes,arrows.meta,backgrounds,fit,calc,decorations.pathmorphing, shapes.geometric, decorations.markings}

\pgfdeclarelayer{middle}
\pgfsetlayers{background,middle,main}
\definecolor{nodeblue}{HTML}{2B6CB0}
\definecolor{nodefill}{HTML}{E8F0FE}
\definecolor{sggreen}{HTML}{276749}
\definecolor{sgfill}{HTML}{E6F4EA}
\definecolor{sensorgray}{HTML}{4A5568}
\definecolor{sensorfill}{HTML}{F7FAFC}
\definecolor{topicamber}{HTML}{C05621}
\definecolor{buildpurple}{HTML}{553C9A}
\definecolor{buildfill}{HTML}{F3F0FF}
\definecolor{searchred}{HTML}{C53030}
\definecolor{searchfill}{HTML}{FFF5F5}
\definecolor{vlmteal}{HTML}{0D7377}
\definecolor{vlmfill}{HTML}{E6FFFA}

\definecolor{bgdark}{HTML}{FFFFFF}        
\definecolor{panelbg}{HTML}{F2F2F6}       
\definecolor{txt}{HTML}{1F2233}           
\definecolor{txtmuted}{HTML}{5A5A6E}      
\definecolor{gnnpurple}{HTML}{6B5BD6}     
\definecolor{scorered}{HTML}{C85454}      
\definecolor{queryorange}{HTML}{E08A2E}   
\definecolor{containerblue}{HTML}{6E7BE0} 
\definecolor{roomorange}{HTML}{E07B4A}    
\definecolor{linearGreen}{HTML}{4F9D6E}   
\definecolor{landmarkGreen}{HTML}{3FA86A} 
\definecolor{normYellow}{HTML}{D9C24A}    
\definecolor{rankedPink}{HTML}{E8F5E8}    
\definecolor{clipOrange}{HTML}{E58843}    
\definecolor{l2blue}{HTML}{6E7BE0}        
\definecolor{forkGrey}{HTML}{6B6B7E}      
\definecolor{skipBlue}{HTML}{5567C9}      
\definecolor{edgegreen}{HTML}{5FBE8E}     
\tikzset{
  font=\sffamily,
  gnngroup/.style={draw=gnnpurple, dash pattern=on 4pt off 3pt, line width=0.9pt,
                   rounded corners=6pt, inner sep=10pt},
  scoregroup/.style={draw=scorered, dash pattern=on 4pt off 3pt, line width=0.9pt,
                     rounded corners=6pt, inner sep=10pt},
  querygroup/.style={draw=queryorange, dash pattern=on 4pt off 3pt, line width=0.9pt,
                     rounded corners=6pt, inner sep=10pt},
  block/.style={rectangle, rounded corners=4pt, draw=none, minimum height=9mm,
                minimum width=26mm, align=center, text=white, font=\sffamily\small},
  pill/.style={rectangle, rounded corners=10pt, draw=none, minimum height=9mm,
               minimum width=26mm, align=center, font=\sffamily\small},
  greenblk/.style={block, fill=linearGreen},
  yellowblk/.style={block, fill=normYellow, text=black},
  orangeblk/.style={block, fill=clipOrange},
  blueblk/.style={block, fill=l2blue},
  containerpill/.style={pill, fill=containerblue!70, text=white, minimum width=30mm},
  roompill/.style={pill, fill=roomorange, text=white, minimum width=22mm},
  rankedpill/.style={pill, fill=rankedPink, text=black, minimum width=34mm, minimum height=11mm},
  forktok/.style={rectangle, rounded corners=3pt, fill=forkGrey, text=white,
                  minimum height=7mm, minimum width=11mm, font=\sffamily\small\itshape},
  roomnode/.style={circle, draw=none, fill=roomorange, text=white,
                   minimum size=12mm, font=\sffamily\small, inner sep=0pt},
  containernode/.style={circle, draw=none, fill=containerblue, text=white,
                        minimum size=10mm, font=\sffamily\scriptsize, inner sep=0pt},
  landmarknode/.style={circle, draw=none, fill=landmarkGreen, text=white,
                       minimum size=10mm, font=\sffamily\scriptsize, inner sep=0pt},
  skipnode/.style={circle, draw=skipBlue, line width=0.8pt, fill=bgdark,
                   text=skipBlue, minimum size=5mm, inner sep=0pt, font=\sffamily\bfseries\small},
  flow/.style={-{Stealth[length=2.3mm]}, line width=1pt, draw=queryorange},
  msgflow/.style={-{Stealth[length=1.8mm]}, line width=0.7pt, draw=linearGreen!90!black},
  rcflow/.style={{Stealth[length=1.8mm]}-{Stealth[length=1.8mm]},
                 line width=1.1pt, draw=roomorange},
  teleport/.style={densely dotted, line width=0.7pt, draw=txtmuted},
  msgpass/.style={dashed, line width=0.7pt, draw=txtmuted},
  innerflow/.style={-{Stealth[length=2mm]}, line width=0.9pt, draw=txt},
}

\definecolor{bg}{HTML}{FFFFFF}
\definecolor{panelbg}{HTML}{EEF1F7}
\definecolor{legendbg}{HTML}{E8ECF3}
\definecolor{textmain}{HTML}{1B2230}
\definecolor{textmuted}{HTML}{5A6478}
\definecolor{cyantext}{HTML}{2A8E91}
\definecolor{greentext}{HTML}{2E9468}
\definecolor{purpletext}{HTML}{6B5BB5}
\definecolor{orangetext}{HTML}{8B6547}
\definecolor{cyanaccent}{HTML}{6FD3D6}
\definecolor{greenaccent}{HTML}{6FD3A3}
\definecolor{purpleaccent}{HTML}{B6A7E6}
\definecolor{orangeaccent}{HTML}{E58C6B}
\definecolor{cardrgb}{HTML}{8DE0E3}
\definecolor{cardtype}{HTML}{6FD3D6}
\definecolor{cardpos}{HTML}{D988A8}
\definecolor{cardroomlabel}{HTML}{52C6CA}
\definecolor{cardltext}{HTML}{3AB8BD}
\definecolor{cardctext}{HTML}{28A8AD}
\definecolor{clipbox}{HTML}{6FD3D6}
\definecolor{spatialbox}{HTML}{E991B3}
\definecolor{visualclip}{HTML}{4D86B5}
\definecolor{textclip}{HTML}{6B8FC7}
\definecolor{sceneclip}{HTML}{4D86B5}
\definecolor{heightcol}{HTML}{A8D498}
\definecolor{proxcol}{HTML}{6B9E5A}
\definecolor{roomtypetxt}{HTML}{A8805A}
\definecolor{meanltext}{HTML}{F0D4B8}
\definecolor{meanctext}{HTML}{B5A88C}
\definecolor{kitchencol}{HTML}{E58C6B}
\definecolor{containercol}{HTML}{9C8FD1}
\definecolor{landmarkcol}{HTML}{4FBE8E}
\definecolor{edgegreen}{HTML}{5FBE8E}
\definecolor{edgeorange}{HTML}{E58C6B}
\definecolor{dashgray}{HTML}{8892A6}

\title{\LARGE \bf
LOCUS: Landmark-Oriented Container Discrimination Using Spatial Graphs
}

\author{Taylor Bergeron$^{*1}$, Shibani Senthilbabu$^{*1}$ and Kevin Leahy$^{1}$
\thanks{$^{*}$These authors contributed equally to this work.}
\thanks{$^{1}$The authors are with the Department of Robotics Engineering, Worcester Polytechnic Institute, Worcester, MA, USA
        {\tt\small \{tabergeron, ssenthilbabu, kleahy\}@wpi.edu}}%
}

\begin{document}
\maketitle
\thispagestyle{empty}
\pagestyle{empty}


\begin{abstract}

As robots are increasingly deployed in unstructured, real-world environments, the ability to reason about complex spatial and semantic relationships among objects remains a fundamental challenge in enabling robust and generalizable manipulation and navigation. For example, deciding where to search for an object that is not in plain sight depends on where it is physically plausible as well as semantically likely. Popular methods such as cosine similarity with CLIP struggle to disambiguate between spatially and semantically similar objects that could contain a target object. We propose Landmark-Oriented Container Discrimination Using Spatial Graphs (LOCUS). To jointly reason about spatial information and semantics, we train a GNN to update CLIP embeddings on a full environment scene graph by passing embedding information between nodes based on proximity. CLIP embeddings, augmented by semantic knowledge from a fusion of household ontologies provide a robust semantic signal. We evaluate in simulation on a noiseless scene graph from simulator metadata, making the approach detection-agnostic. Our approach outperforms random, CLIP, Tidybot, and an LLM planner in the majority of room classes and configurations in simulation. To show our approach is robust to scene graph node placement and label noise, we demonstrate a physical mobile manipulator running in an exploration pipeline start-to-finish including scene graph labels from Detic.

\end{abstract}



\section{Introduction}

To accomplish many complex tasks in the real world, robots must understand physical and semantic relationships between objects in their environment. For example, to find occluded objects in household or search-and-rescue tasks, a robot needs both general and domain-specific knowledge of object relationships to hypothesize where an object may be hidden, including inside other containers.
The robot may also be provided with a general floor plan which it can use to inform its search, but have no knowledge about the items within each room. In many cases, there are multiple instances of the same container object which could contain the target object, such as multiple drawers that could contain a fork, and the robot must find a way to discriminate between the containers to generate a container search order. We posit that the correct approach requires both the generality of Large Language Models (LLMs) or Vision Language Models (VLMs) and the structured knowledge of an ontology.

We introduce a novel approach, Landmark-Oriented Container discrimination Using Spatial Graphs (LOCUS), the first reasoning pipeline for concealed objects that makes instance-level decisions between containers of the same class. Locus is Latin for place — reflecting our hypothesis that where a container is matters as much as what it looks like. LOCUS combines domain specific formal ontologies and general semantic knowledge. It accounts for cases not explicitly covered in the ontologies with the spatial relationship between objects in the environment in order to find an occluded object and is generalizable sim-to-sim and sim-to-real. We pretrain a GNN over scene graphs and use ontologies to learn typical spatial relationships and \textit{augment} each scene graph node's CLIP ~\cite{radfordLearningTransferableVisual2021} embedding.


Our contributions include:
\begin{inparaenum}
    \item a formal ontology for sorting objects into containers, landmarks, or dual role;
    \item a dataset with ontology-informed placement strategy for occluded objects in containers (\href{https://dataverse.harvard.edu/previewurl.xhtml?token=7d65872b-7a2a-42e5-a52a-828f38c8d542}{\textit{iTHOR Object-Container Spatial Relationship Dataset for LOCUS}});
    \item a scene graph builder that uses these ontologies; and
    \item a room-agnostic GNN architecture to be trained on explored environment scene graph that accurately passes semantic information from the environment to update CLIP embeddings. We will publish our codebase upon acceptance (\href{https://dataverse.harvard.edu/previewurl.xhtml?token=17c12ead-2431-4f92-b1b3-9645fd70611a}{\textit{LOCUS Codebase}}).
\end{inparaenum}

\section{Background and Problem Formulation}

To reason over object search and retrieval solutions, many approaches ~\cite{maggio_clio_2024, yokoyama2024vlfm, gadreCoWsPastureBaselines2023} use CLIP for retrieval, comparing search locations by cosine similarity with the query; however, same-class containers receive nearly identical embeddings, whether from a bounding-box crop or a wider frame containing context (e.g., a drawer with the sink visible). Methods exist for semantic object search ~\cite{zhouESCExplorationSoft2023, shahNavigationLargeLanguage2023, dorbalaCanEmbodiedAgent2024, wangOVAMOSFrameworkOpenVocabulary2025, zhaoIntegratingSemanticPriors2025, gintingSEEKSemanticReasoning2024, kurenkovSemanticGeometricModeling2021} and container reasoning ~\cite{mirakhorTaskPlanningObject2024, kant2022housekeep, sarchTIDEETidyingNovel2022}, but none produce a principled pipeline to search for an occluded object in a realistic environment with multiple containers of the same class to choose between during exploration given a floor plan and an ontology for the domain. Scene-graph GNNs. rank navigation goals in observable space~\cite{lmnav, hron}; HOLM~\cite{holm} and VLMaps~\cite{huangVisualLanguageMaps2023} resolve only viewpoint occlusion, where the target becomes visible once the agent looks in the right direction; semantic mechanical search~\cite{sharma2023semanticmechanicalsearchlarge} reasons over occupancy rather than which of several identical containers to open.

\textbf{Semantic World Knowledge for Object Search Decisions} Many approaches investigate commonsense reasoning while exploring for not completely occluded objects, with LLMs as a popular solution. The approaches that reason over room-object navigation decisions for \textit{unknown environments} either combine preexisting pretrained models for image parsing and an LLM to reason over objects in the image to inform the search ~\cite{zhouESCExplorationSoft2023, shahNavigationLargeLanguage2023}, use LLM chain-of-thought prompting for commonsense context ~\cite{dorbalaCanEmbodiedAgent2024}, or build multi-layer VLM value maps from LLM-inferred semantic priors ~\cite{wangOVAMOSFrameworkOpenVocabulary2025, zhaoIntegratingSemanticPriors2025}. An approach to incorporating environment understanding into the reasoning pipeline is using a scene graph of the current environment to aid in decision making. For \textit{partially known environments} with a floor plan, ~\cite{gintingSEEKSemanticReasoning2024} and ~\cite{kurenkovSemanticGeometricModeling2021} use a scene graph to inform their neural network based approaches. ~\cite{gintingSEEKSemanticReasoning2024} use LLMs to generate training data for their network which can lead to hallucinations and ~\cite{kurenkovSemanticGeometricModeling2021} manually craft the semantic categories governing where objects are placed, which makes the vocabulary sparse and unable to accurately reflect or scale to real-world distributions. These approaches lack principled semantic grounding through ontologies or human placement data, which would impact performance for occluded object search.

\textbf{Container Reasoning} Approaches for object-container reasoning don't consider instances of containers but rather consider object to container class relationship (e.g. a fork goes in a drawer, and all drawers are treated as the same since they belong to the same class) ~\cite{mirakhorTaskPlanningObject2024, kant2022housekeep, sarchTIDEETidyingNovel2022}. Approaches that consider finding entirely occluded objects rely on simulation metadata at runtime and abstract away manipulation task choices, as if the agent is within range of a fully occluded object, then it is found ~\cite{mirakhorTaskPlanningObject2024, kant2022housekeep}, while the approach that considers sensor data for labels using d-DETR ~\cite{zhu2021ddetr} rather than the ground truth labels from simulation metadata at run time only considers fully visible objects ~\cite{sarchTIDEETidyingNovel2022}. TidyBot~\cite{tidybot} likewise selects a receptacle \emph{type} from LLM preferences, so it cannot order same-class instances. To our knowledge, no prior work discriminates same-class container instances under full occlusion. Notably, all of these approaches are in simulation and do not generalize cross-simulation. There is a gap for real world robotics deployment when the robot needs to choose between containers, some of which are of the same class, to search to physically retrieve an occluded object.
Therefore, our goal is to answer, \emph{Given a room level floor plan of an environment and task/environment relevant ontologies, can a robot distinguish between instances of the same object class with different semantic and spatial information and generalize to unseen environment configurations and domains?}

Our work focuses on bridging the gap between semantic world knowledge for object search decisions and instance container reasoning for realistic object retrieval. CLIP provides strong semantic understanding. Rather than training an entirely new network, we update CLIP with spatial context for instance-level container discrimination via a GNN that learns over an environment's scene graph.

To showcase LOCUS's ability to provide detailed distinctions between nodes of the same class, at run time we score nodes by cosine similarity between the updated container embedding and the query text, retaining CLIP's comparability while adding spatial and semantic reasoning.


\section{Technical Approach}





\subsection{Graph Construction and Ontology}
\label{subsec:graph_construction_and_ontology}

\begin{figure*}
    \centering
    \resizebox{\linewidth}{!}{\input{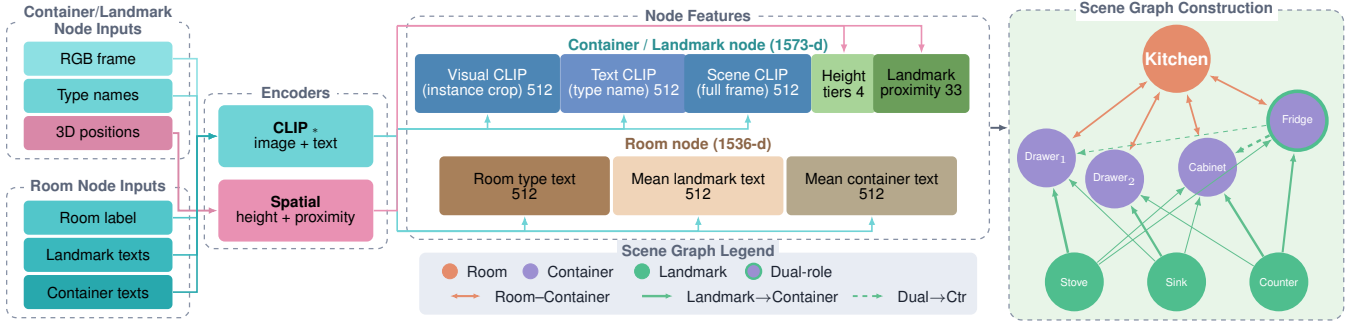}}
    \caption{Feature extraction and graph construction. }
    \label{fig:pipeline_construction}
\end{figure*}

LOCUS represents a scene as a heterogeneous \emph{scene graph} $\mathcal{G} = (\mathcal{V}, \mathcal{E})$ where each detected object in the environment is a node $v \in \mathcal{V}$ and edges $e \in \mathcal{E}$ encode spatial proximity between objects. To construct the graph, as shown in Fig.~\ref{fig:pipeline_construction}, every object must be assigned a \emph{node role} that determines how it participates in reasoning. Prior scene-graph methods assign such roles through hierarchical graph layers~\cite{werbyHierarchicalOpenVocabulary3D2024, kurenkovModelingDynamicEnvironments, mahdiRelationalSemanticReasoning2026} or task-specific heuristics~\cite{zengSemanticLinkingMaps2020}. These categorizations do not generalize: they provide no criterion for classifying a newly encountered object. We instead derive node roles from formal ontologies (see Sec. \ref{subsec:ontology}).

We define three node roles. \emph{Containers} are receptacles that can be opened (e.g. drawers, cabinets, refrigerators) and are the objects the robot must rank when searching for a hidden target. \emph{Landmarks} are large, stationary objects that cannot be picked up or opened (e.g., countertops, stove burners, sinks) and serve as spatial reference points. \emph{Dual-role} nodes are containers that also function as landmarks: a refrigerator is both a searchable receptacle and a spatial anchor that contextualizes nearby drawers and cabinets. In addition to these object nodes, a single \emph{room} node is added to represent the room class (e.g., kitchen, bathroom) and provide room-level semantic context.

Two types of edges connect nodes in the graph. \emph{Landmark-to-container} edges are directed from each landmark to nearby containers, weighted by $w = \exp(-d)$ where $d$ is the Euclidean distance, so closer landmarks exert stronger influence during message passing. \emph{Room-to-container} edges are bidirectional and connect the room node to every container with unit weight, allowing room-class context to flow to containers and aggregated container information to flow back to the room node.

\subsection{Graph Neural Network}
\label{subsec:gnn}
\begin{figure*}
    \centering
     \resizebox{.8\linewidth}{!}{\input{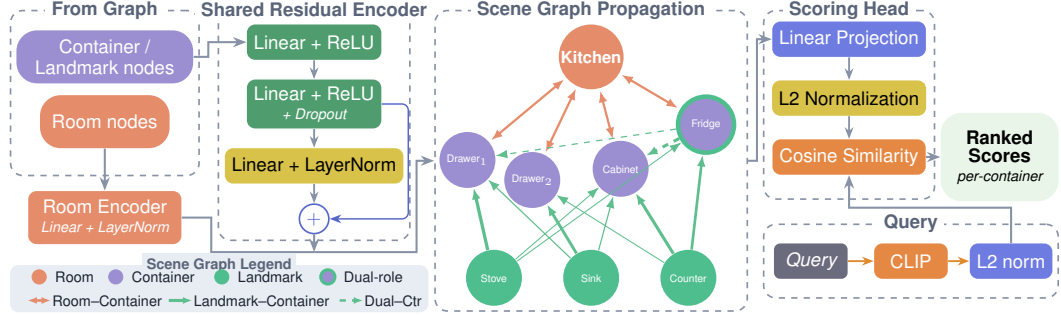}}
    \caption{GNN architecture and scoring. A shared residual encoder maps container and landmark features; APPNP propagates over the graph.}
    \label{fig:pipeline_gnn}
\end{figure*}

We require a mechanism to propagate spatial and semantic context between scene graph nodes so that containers of the same class, with nearly identical CLIP embeddings, can be distinguished by their neighborhood. We use a graph neural network (GNN) to update each node's representation by aggregating information from connected landmarks and the room node, as shown in Fig.~\ref{fig:pipeline_gnn}.

\subsubsection{Semantic Information}
\label{subsubsec:semantic}
To distinguish between nodes of the same class, a node's representation must capture not only what the object looks like but also its location relative to the room and other object nodes. Each node's feature vector contains visual appearance, semantic identity, spatial context, and relative position.

Each container node concatenates a CLIP visual embedding of the container's bounding box crop, aggregated across all viewpoints by element-wise maximum to retain the most distinctive features from any angle (maxpooling); a CLIP text embedding of the container's class label; a full-frame CLIP embedding from the viewpoint where the container occupies the largest pixel area, capturing surrounding scene context; and a height encoding that bins vertical position into floor, counter, and wall tiers (a 4-d feature: normalized height plus the three-tier one-hot). A soft landmark-distance vector is appended, encoding proximity to each of the landmark types observed during training as $\exp(-d)$ for the nearest instance of each type.
At deployment, hard class assignments are replaced by soft assignments: each detection's CLIP embedding is compared against per-class prototype embeddings learned from training, and a softmax produces a weight vector over landmark types. Landmark nodes use the same input layout as well. Room nodes concatenate the CLIP text embedding of the room class with the mean text embeddings of member containers and landmarks. All container and landmark features are mapped to a common embedding space through a shared residual encoder. Room nodes use a separate linear projection.

\subsubsection{Message Passing}
\label{subsubsec:message}
After encoding, each node's representation reflects only its own features. Message passing allows containers to absorb context from their spatial neighbors: a drawer near a sink should be represented differently from a drawer near a stove, even if both look identical.

We use APPNP~\cite{gasteiger2019appnp}  to propagate information across the graph. Edges run from landmarks to containers with weight $\exp(-d)$, and from rooms to containers with unit weight. After propagation, container embeddings are projected through a linear head, L2-normalized, and scored against query text embeddings by cosine similarity.

\subsubsection{Loss}
\label{subsubsec:loss}

The primary objective combines two contrastive losses. The first loss is a multi-positive InfoNCE ~\cite{oord2018representation} that, for each query, contrasts every container holding the target object against all containers and landmarks in the scene:
\begin{equation}
  \mathcal{L}_{\text{InfoNCE}}
    = -\frac{1}{|\mathcal{P}|}\sum_{i \in \mathcal{P}}
      \log \frac{\exp(s_i / \tau)}
           {\sum_{j \in \mathcal{C} \cup \mathcal{L}}
            \exp(s_j / \tau)}
\label{eq:infonce}
\end{equation}
where $s_k = \cos(\mathbf{z}_k, \mathbf{q})$ is the cosine similarity between node $k$'s post-GNN embedding and the query text embedding, with the temperature $\tau$ applied only inside the exponent, $\mathcal{P}$ is the set of positive containers, $\mathcal{C}$ all containers, and $\mathcal{L}$ all landmarks in the scene.
Landmarks are unlabeled: they appear in the denominator but never receive positive or negative labels. Each query is weighted by its inverse positive rate across the training set.

The second loss is a soft ranking loss that penalizes the expected number of containers a robot would open before finding the target:
\begin{equation}
  \mathcal{L}_{\text{search}}
    = \frac{1}{|\mathcal{P}|}\sum_{i \in \mathcal{P}}
      \log\!\Bigl(1 + \sum_{j \in \mathcal{C} \cup \mathcal{L}}
        \sigma(s_j - s_i)\Bigr)
\label{eq:searchfull}
\end{equation}
where $\sigma$ is the sigmoid function at unit temperature. This loss directly targets our primary evaluation metric Expected Containers Opened (ECO) (see Sec. \ref{baselinesandmetrics}.)

Two auxiliary losses regularize the encoder during training
. A binary cross-entropy head, $\mathcal{L}_{\text{contain}}$, on stop-gradient embeddings predicts $\hat{y} = \sigma(f(\mathbf{z}))$, the probability that node $\mathbf{z}$ is a container. A learned scalar $\beta$ boosts container scores by $+\beta\,\hat{y}$ and suppresses landmarks by $+\beta\,(\hat{y} - 1)$. A second auxiliary head,$\mathcal{L}_{\text{teacher}}$, matches each container's representation to a soft K-means cluster assignment derived from open-view CLIP embeddings, preventing the encoder from collapsing visual distinctions that closed-door appearance obscures.

The total loss is as follows, where $\lambda_{sf}$,$ \lambda_c$, and $\lambda_t$ are constants used to weight the respective losses:
\begin{equation}
  \mathcal{L}
    = \mathcal{L}_{\text{InfoNCE}}
    + \lambda_{\text{sf}}\,\mathcal{L}_{\text{search}}
    + \lambda_{\text{c}}\,\mathcal{L}_{\text{contain}}
    + \lambda_{\text{t}}\,\mathcal{L}_{\text{teacher}}
\label{eq:total_loss}
\end{equation} 


\section{Experiment}
\label{sec:result}

\subsection{Task}

We task the robot to search for an item given its text label (e.g., fork). The item may be visible (found through mapping to create the scene graph) or entirely occluded within a container object, in which case the robot must decide which containers to search and in what order.

\subsection{Ontology}
\label{subsec:ontology}

The ontology is used only for dataset creation. We define three node roles using simulator metadata and formal knowledge bases in Sec. ~\ref{subsec:graph_construction_and_ontology}. \emph{Containers} are objects flagged as both \texttt{openable} and \texttt{receptacle} in AI2-THOR~\cite{kolveAI2THORInteractive3D2022}. \emph{Landmarks} are non-pickupable, non-openable objects classified under landmark-relevant upper classes in at least one of three ontological backends: SUMO~\cite{niles2001sumo} (\texttt{Furniture}, \texttt{MajorAppliance}, \texttt{ElectricDevice}, \texttt{CoolingDevice}, etc.), SOMA~\cite{bessler2021soma} (\texttt{DesignedFurniture}, \texttt{Appliance}, \texttt{DesignedComponent}), or WordNet (\texttt{furnishing.n.02}, \texttt{appliance.n.02}, \texttt{fixture.n.01}). Countertops and stove burners qualify as landmarks while garbage bags do not.

\subsection{Metrics}

We report four metrics. \textbf{Instance Rank Accuracy (IRA)} measures within-class discrimination: among all containers of the same class, is the top-scored instance the ground-truth positive? IRA is computed only over \emph{discriminative} queries---those where at least one same-class container holds the target and at least one does not. \textbf{Expected Containers Opened (ECO)} is the rank of the first positive container in descending score order---the number of containers a robot would open before finding the target. \textbf{ECO/$N$} normalizes ECO by the number of containers $N$ in each scene, giving the fraction of containers opened before finding the target (0~=~first try, 1~=~every container). This enables direct comparison across environments with different container counts (e.g., iTHOR~${\sim}10$ vs.\ RoboCasa~${\sim}34$). \textbf{S-IRA} (Semantic IRA) relaxes IRA so that any co-placement equivalent container of the same class counts as correct, accounting for functional interchangeability (e.g., two drawers that both hold forks). \textbf{S-ECO/$N$} (Semantic ECO/$N$) applies the same relaxation to ECO/$N$. We report both because their gap is itself the measurement: the ECO/$N$ vs S-ECO/$N$ gap quantifies the errors that select a functionally equivalent container. IRA${=}1.0$ is unattainable under partial information, as identical drawers cannot be told apart without priors; therefore ECO is our primary metric, and it also covers more of the test set. The 42 query classes yield 201 (scene, query) episodes, of which ECO uses all 201 and IRA only the 165 discriminative ones (hence dashes in Table~\ref{tab:perquery}).

\subsection{Baselines}
\label{baselinesandmetrics}

\textbf{Random} selects containers uniformly (5-seed average). \textbf{Visual CLIP}~\cite{radfordLearningTransferableVisual2021} ranks containers by cosine similarity between the query text embedding and the full-frame ViT-B/32 image embedding of the frame in which each container was most visible. \textbf{Class Text CLIP} ranks by cosine similarity between the query text and the CLIP text embedding of each container's class name. \textbf{Priors} rank by $P(q{\mid}\text{room}){\times}P(q{\mid}\text{type})$, optionally scaled by a landmark-distance term $f(\text{dist}){=}e^{-d/\sigma}$. \textbf{TidyBot}~\cite{tidybot} is ported into our evaluation pipeline. The commonsense (zero-shot) and summarization (informed with our training placements) variants select a receptacle type, per their design, and instances within type are randomly selected (chance IRA, so this upper bound covers ConceptGraphs too ~\cite{guConceptGraphsOpenVocabulary3D2024}). \textbf{LLM scene graph} gives an LLM (Claude Haiku 4.5) each test scene's full serialized graph
and requests a ranking over container instances zero-shot. \textbf{LLM raw detections} is the ontology-free and graph-free counterpart, so the same LLM model ranks instances from the raw detection list. \textbf{MLP over CLIP} with full spatial features, is the toggled $K{=}0$.


\section{Results}

\subsection{iTHOR Simulation}

\subsubsection{Setup}
\label{sec:dataset}

 Training and evaluation use iTHOR to isolate the GNN from perception noise. The scene graph comes directly from simulator metadata, so there is no exploration phase in simulation or sim-to-sim transfer.

\textbf{Dataset} We demonstrate LOCUS in the household domain, due to the multiple household ontologies~\cite{niles2001sumo, bessler2021soma}.
We collect data from all 120 iTHOR~\cite{kolveAI2THORInteractive3D2022} scenes spanning four room classes (30 kitchens, 30 living rooms, 30 bedrooms, 30 bathrooms), split into 80 train / 20 validation / 20 test scenes stratified by room class. For each scene an exhaustive grid sweep captures RGB frames at every reachable position, four yaw rotations, and three camera tilts, producing dense multi-view coverage.

AI2-THOR's default scenes have mostly empty containers, so we define a semantic placement policy that distributes 42 query-object classes into openable receptacles. The class-level mappings follow the same conventions validated by HouseKeep's~\cite{kant2022housekeep} crowdsourced annotations (e.g.\ cutlery in Drawer, Cabinet in kitchens; perishables in Fridge or Cabinet). Objects that serve the same purpose, such as fork or spoon, are placed together into the same container when capacity allows, and containers near a semantically related landmark are preferred (e.g.\ pots near a stove burner). A soft capacity limit of five objects per container prevents unrealistic crowding. Three random seeds produce independent placements per scene; ground-truth labels are aggregated across seeds, yielding 99.6\% cross-seed agreement.
Our placement rules additionally create spatial variation at the instance level; two drawers in the same kitchen may hold different objects depending on their proximity to landmarks, which is the discrimination signal LOCUS is designed to learn.

\textbf{Training} We train on the 80-scene training split above for 500 epochs with learning rate $10^{-4}$, weight decay $10^{-4}$, batch size 32, and $\tau\!=\!0.07$. The encoder is a 3-layer residual MLP with skip connections and LayerNorm. APPNP~\cite{gasteiger2019appnp} uses $K\!=\!2$ hops and teleport $\alpha\!=\!0.3$. We evaluate on a held-out 20-scene test split (5 per room class) and average all LOCUS results over 30 training seeds. For training we assume the robot has perfect sensor readings of the environment, so it has correct object class labels and poses during the training process. 

We do not train with perception noise or Detic labels: the GNN consumes only the scene graph from simulator metadata, and Detic labels are noisy pseudo-labels. The zero-shot Stretch~3 run with real perception error provides that evidence instead(See Section \ref{mobilemanipulatorresults}).

\subsubsection{Results}

Table~\ref{tab:main} reports all metrics on the 201-episode iTHOR test split. TidyBot, ConceptGraphs~\cite{guConceptGraphsOpenVocabulary3D2024}, and the $P(q{\mid}\text{room}){\times}P(q{\mid}\text{type})$ prior score containers by class so the first container opened is a random pick among all the containers sharing that class. This means the upper bound on IRA for all these methods is uniform random chance, regardless of how accurate the class prediction is. Adding $f(\text{dist})$ to the prior only raises IRA because its distance term matches the rule we used to place objects, so that gain is circular, and it still loses every search-cost metric. Both CLIP baselines perform at or below random on IRA, unable to distinguish same-class instances. Given raw detections alone the LLM names the correct class at nearly LOCUS's rate, yet its IRA is close to random, which supports our claim of instance discrimination being the bottleneck. ECO on the other hand does not have that upper bound, yet we outperform the baselines. TidyBot, which names the correct class 92.0\% of the time yet opens .338$N$ containers per search, against LOCUS which names the correct class 95.2\% and opens.257$N$ per search, a 24.0\% improvement. 
A stronger LLM model baseline would likely not raise the ECO score, since the missing information is about instances, not classes. 
The LLM can reason about instances as it is given the full scene graph, so it is the strongest baseline on ECO search cost but trails LOCUS on every metric.

The $K\!=\!0$ ablation, an MLP over the same features with no message passing, already reaches an IRA 40\% higher than the chance upper bound for methods ranking containers by class alone, showing that the spatial features alone are enough to reason about instances rather than classes. APPNP propagation ($K\!=\!2$) increases IRA by another 7\% confirming that inter-node context provides a modest but consistent benefit across all 30 training seeds ($p{<}.001$, paired $t$). \textbf{S-ECO/$N$} counts functionally equivalent containers as successes, and doing so lowers search cost by 17\% (ECO 2.83 to S-ECO 2.17 containers), so much of what ECO determines as error is LOCUS opening a container that would also have been a reasonable place to look.


\begin{table}[t]
\centering
\caption{iTHOR test set. Best result per column in \textbf{bold}. Informed=our training placements.
}
\label{tab:main}
\small
\setlength{\tabcolsep}{2.5pt}
\resizebox{\linewidth}{!}{%
\begin{tabular}{|l|c|c|c|c|}
\hline
\textbf{Method} & \textbf{IRA}\,$\uparrow$ & \textbf{ECO}/$N$\,$\downarrow$ & \textbf{S-ECO}/$N$\,$\downarrow$ & \textbf{TypeAcc}\,$\uparrow$ \\
\hline
Random (5 seeds)              & .241 & .455 & .347 & 63.3\% \\
Visual CLIP                   & .188 & .466 & .375 & 60.2\% \\
Text CLIP                     & .182 & .436 & .385 & 52.2\% \\
\hline
$P(q{\mid}\text{room}){\times}P(q{\mid}\text{type})$ & .152 & .398 & .356 & 61.7\% \\
$P(q{\mid}\text{room}){\times}P(q{\mid}\text{type}){\times}f(\text{dist})$ & \textbf{.455} & .335 & .288 & 60.7\% \\
\hline
TidyBot commonsense (zero-shot)  & .225${\pm}$.019 & .338\,$\pm$\,.008 & .277\,$\pm$\,.007 & 92.0\% \\
TidyBot summarization (informed)  & .225${\pm}$.019 & .332\,$\pm$\,.007 & .273\,$\pm$\,.007 & 89.6\% \\
LLM raw detections (Haiku 4.5) & .236 & .317 & .259 & 95.0\% \\
LLM scene graph (Haiku 4.5)   & .279 & .284 & .245 & 88.6\% \\
\hline
LOCUS ($K\!=\!0$) (MLP) & .342\,$\pm$\,.023 & .272\,$\pm$\,.014 & .223\,$\pm$\,.010 & 93.4\%\\
\textbf{LOCUS} ($K{=}2$) & .367\,$\pm$\,.024 & \textbf{.257\,$\pm$\,.012} & \textbf{.212\,$\pm$\,.009} & \textbf{95.2\%} \\
\hline
\end{tabular}%
}
\end{table}

\begin{table}[t]
\small
\centering
\caption{Per-room breakdown IRA and ECO. $n$ = episodes per room (201 total). mean$\pm$std over 30 seeds, LOCUS: training seeds, baseline: tiebreak seeds.}
\label{tab:perroom}
\resizebox{\linewidth}{!}{%
\begin{tabular}{|l|c|l|c|c|c|c|}
\hline
\textbf{Room} & $\mathbf{n}$ & \textbf{Method} & \textbf{IRA} $\uparrow$ & \textbf{S-IRA} $\uparrow$ & \textbf{ECO/$N$} $\downarrow$ & \textbf{S-ECO/$N$} $\downarrow$\\
\hline
Bath. & 32  & Random      & .383{\scriptsize$\pm$.074} & .554{\scriptsize$\pm$.068} & .498{\scriptsize$\pm$.044} & .423{\scriptsize$\pm$.040} \\
      &     & Vis.\ CLIP  & .414{\scriptsize$\pm$.000} & .586{\scriptsize$\pm$.000} & .478{\scriptsize$\pm$.000} & .403{\scriptsize$\pm$.000} \\
      &     & Text CLIP   & .382{\scriptsize$\pm$.071} & .555{\scriptsize$\pm$.068} & .547{\scriptsize$\pm$.023} & .493{\scriptsize$\pm$.018} \\
      &     & LOCUS       & \textbf{.653}{\scriptsize$\pm$.063} & \textbf{.678}{\scriptsize$\pm$.048} & \textbf{.306}{\scriptsize$\pm$.011} & \textbf{.302}{\scriptsize$\pm$.009} \\
\hline
Kit.  & 108 & Random      & .171{\scriptsize$\pm$.034} & .312{\scriptsize$\pm$.038} & .434{\scriptsize$\pm$.022} & .292{\scriptsize$\pm$.016} \\
      &     & Vis.\ CLIP  & .065{\scriptsize$\pm$.006} & .108{\scriptsize$\pm$.009} & .458{\scriptsize$\pm$.001} & .336{\scriptsize$\pm$.001} \\
      &     & Text CLIP   & .166{\scriptsize$\pm$.033} & .306{\scriptsize$\pm$.035} & .348{\scriptsize$\pm$.011} & .291{\scriptsize$\pm$.009} \\
      &     & LOCUS       & \textbf{.283}{\scriptsize$\pm$.042} & \textbf{.440}{\scriptsize$\pm$.049} & \textbf{.166}{\scriptsize$\pm$.018} & \textbf{.108}{\scriptsize$\pm$.011} \\
\hline
Liv.  & 23  & Random      & .312{\scriptsize$\pm$.093} & .327{\scriptsize$\pm$.094} & .546{\scriptsize$\pm$.045} & .529{\scriptsize$\pm$.044} \\
      &     & Vis.\ CLIP  & \textbf{.400}{\scriptsize$\pm$.000} & \textbf{.400}{\scriptsize$\pm$.000} & .542{\scriptsize$\pm$.004} & .531{\scriptsize$\pm$.004} \\
      &     & Text CLIP   & .295{\scriptsize$\pm$.095} & .308{\scriptsize$\pm$.095} & .620{\scriptsize$\pm$.039} & .603{\scriptsize$\pm$.038} \\
      &     & LOCUS       & .368{\scriptsize$\pm$.053} & .395{\scriptsize$\pm$.059} & \textbf{.471}{\scriptsize$\pm$.022} & \textbf{.464}{\scriptsize$\pm$.021} \\
\hline
Bed.  & 38  & Random      & .236{\scriptsize$\pm$.082} & .366{\scriptsize$\pm$.088} & .420{\scriptsize$\pm$.035} & .342{\scriptsize$\pm$.027} \\
      &     & Vis.\ CLIP  & .188{\scriptsize$\pm$.023} & .308{\scriptsize$\pm$.033} & .424{\scriptsize$\pm$.007} & .366{\scriptsize$\pm$.005} \\
      &     & Text CLIP   & .216{\scriptsize$\pm$.079} & .345{\scriptsize$\pm$.092} & .423{\scriptsize$\pm$.034} & .349{\scriptsize$\pm$.027} \\
      &     & LOCUS       & \textbf{.327}{\scriptsize$\pm$.057} & \textbf{.459}{\scriptsize$\pm$.066} & \textbf{.345}{\scriptsize$\pm$.024} & \textbf{.280}{\scriptsize$\pm$.023} \\
\hline
\end{tabular}}
\end{table}

\begin{table}
\centering
\caption{Per-query IRA and ECO. $n$ = episodes per query.}
\label{tab:perquery}
\begin{tabular}{|l|r|cc|cc|}
\hline
& & \multicolumn{2}{c|}{\textbf{Vis.\ CLIP}} & \multicolumn{2}{c|}{\textbf{LOCUS}} \\
\textbf{Query} & $n$ & IRA $\uparrow$ & ECO $\downarrow$ & IRA $\uparrow$ & ECO $\downarrow$\\
\hline
fork & 5 & .00 & 16.8 & \textbf{.43} & \textbf{4.5} \\
knife & 5 & .00 & 16.4 & \textbf{.25} & \textbf{3.4} \\
spoon & 5 & .00 & 14.2 & \textbf{.47} & \textbf{3.0} \\
butterknife & 5 & .00 & 17.4 & \textbf{.48} & \textbf{4.0} \\
spatula & 5 & .00 & 13.6 & \textbf{.41} & \textbf{5.1} \\
\hline
cup & 5 & .00 & 7.8 & \textbf{.31} & \textbf{2.9} \\
mug & 5 & .20 & 8.2 & \textbf{.41} & \textbf{2.9} \\
bowl & 5 & .00 & 9.2 & \textbf{.35} & \textbf{4.2} \\
plate & 5 & .20 & 5.0 & \textbf{.35} & \textbf{3.0} \\
\hline
apple       & 5 &  -- &  6.8 &  -- & \textbf{1.0} \\
tomato      & 5 &  -- &  4.8 &  -- & \textbf{1.0} \\
bread       & 5 &  -- & 11.4 &  -- & \textbf{1.0} \\
\hline
creditcard & 10 & .25 & 3.5 & \textbf{.36} & \textbf{2.3} \\
keychain & 10 & .37 & 2.4 & .36 & \textbf{2.1} \\
cell phone & 6 & .00 & 4.0 & \textbf{.17} & \textbf{2.7} \\

\hline
\end{tabular}
\end{table}

Tables~\ref{tab:perroom} and~\ref{tab:perquery} break down results by room class and query object. Kitchens are hardest, yet LOCUS achieves ECO/$N$ 63\% better than random and S-ECO/$N$ less than ECO/$N$, meaning over a third of kitchen errors are semantically equivalent. Living rooms are worst (ECO/$N\!=\!0.471$) because fewer landmarks result in less spatial signal to exploit. Per-query, LOCUS dominates utensils and tableware where spatial context resolves which drawer to open, while fridge-stored foods are trivial for any method with container-class priors (ECO$\!=\!1.0$). IRA .367 means the correct instance is ranked first ${\sim}37\%$ of the time; therefore, LOCUS reduces search cost but it does not reliably identify the target on the first opening.

\textbf{Failures} LOCUS errors are less costly. Errors are right-type wrong-instance or wrong-type. Low IRA does not limit usefulness as 90\% of LOCUS errors are right-type wrong-instance, so adding a few openings rather than failing the search, while ${\sim}50\%$ of baseline errors are costly wrong-type (derived from Table~\ref{tab:typeacc}; per-room breakdown in Table~\ref{tab:error_taxonomy}). 

\textbf{Compute} The average iTHOR scene graph has 10 containers, 11 landmarks, and 122 edges. The fully-connected graph matches the best tuned cutoff and handles RoboCasa's 20--40 containers.

TidyBot requires an LLM call per query at run time, with occasional unparseable output that our port handles fairly by tying remaining candidates. LOCUS is 3.18M parameters at ${<}3$~ms on CPU, offline and deterministic. 

\begin{table}[h]
\centering
\small
\setlength{\tabcolsep}{4pt}
\caption{Class-level against instance-level accuracy, iTHOR test set. 
Superscripts are two-sided paired $t$-tests on per-query values against LOCUS ($K{=}2$) ($^{***}p{<}.001$, $^{**}p{<}.01$, $^{*}p{<}.05$, $^{\text{ns}}$ otherwise)
}
\label{tab:typeacc}
\hspace*{-10pt}%
\begin{tabular}{|l|c|c|c|}
\hline
\textbf{Method} & \textbf{Class} (\%)~$\uparrow$ & \textbf{Inst.} (\%)~$\uparrow$ & \textbf{Inst./Class}~$\uparrow$ \\
\hline
\textit{Constant: cabinets first} & \textit{74.6}$^{*}$ & \textit{21.9$\pm$2.4}$^{***}$ & \textit{29\%} \\
\hline
Random                            & 63.3$\pm$2.6$^{***}$ & 20.1$\pm$2.6$^{***}$ & 32\% \\
Visual CLIP                       & 60.2$^{***}$ & 18.4$^{***}$ & 31\% \\
Class Text CLIP                   & 52.2$^{***}$ & 20.9$^{***}$ & 40\% \\
\hline
TidyBot commonsense               & 92.0$^{\text{ns}}$ & 33.4$\pm$2.6$^{***}$ & 36\% \\
TidyBot summarization             & 89.6$^{\text{ns}}$ & 34.0$\pm$2.4$^{***}$ & 38\% \\
LLM scene graph                   & 88.6$^{\text{ns}}$ & 37.3$^{**}$ & 42\% \\
\hline
LOCUS ($K\!=\!0$)                 & 93.4$\pm$4.7 & 47.1$\pm$2.2 & 50\% \\
\textbf{LOCUS} ($K\!=\!2$)        & \textbf{95.2$\pm$4.8} & \textbf{48.6$\pm$2.5} & \textbf{51\%} \\
\hline
\end{tabular}
\end{table}

\begin{table}[h]
\centering
\small
\caption{Error taxonomy across 201 test episodes (mean\,$\pm$\,std, 30 seeds).}
\label{tab:error_taxonomy}
\begin{tabular}{|l|c|c|c|}
\hline
\textbf{Room} & \textbf{Correct} & \textbf{\makecell{Right type,\\wrong inst.}} & \textbf{Wrong type} \\
\hline
Overall     & 48.6\,$\pm$\,2.5 & 44.6\,$\pm$\,3.9 & 6.8\,$\pm$\,4.8 \\
Kitchen     & 47.3\,$\pm$\,4.2 & 43.0\,$\pm$\,7.1 & 9.7\,$\pm$\,8.8 \\
Living room & 40.7\,$\pm$\,4.6 & 50.7\,$\pm$\,4.5 & 8.6\,$\pm$\,0.8 \\
Bedroom     & 43.3\,$\pm$\,4.8 & 56.7\,$\pm$\,4.8 & 0.0\,$\pm$\,0.0 \\
Bathroom    & 65.0\,$\pm$\,6.3 & 31.4\,$\pm$\,5.6 & 3.6\,$\pm$\,1.7 \\
\hline
\end{tabular}
\end{table}

\subsubsection{Circularity and Determinism}

To rule out the possibility of circularity, in which LOCUS would learn the label generator rather than spatial conventions, we tested relabeled the test split to strip out both of our generator's contributions: the landmark-proximity rule that chose which instance, replaced by a uniform draw, and our type conventions, replaced by HouseKeep~\cite{kant2022housekeep} crowd votes (Table~\ref{tab:hk_labels}). A model that had memorized the generator would lose accuracy in proportion, but zero-shot on these labels, LOCUS's search cost, ECO/$N$, is essentially unchanged. HouseKeep's votes place 89\% of episodes in a cabinet, so the margin to our comparators narrows and a query-independent ranker that opens cabinets first is best under HouseKeep labels and near-worst under ours.



\textit{\underline{Determinism:}} 99.6\% measures the generator's cross-seed reproducibility. 
The near-equal RoboCasa performance on novel and trained queries  (Table~\ref{tab:robocasa_dist}) suggest that LOCUS is learning neither the script's rules nor specific object--container pairings.

\begin{table}[h]
\centering
\small
\setlength{\tabcolsep}{3pt}
\caption{ECO/$N$ and TypeAcc from both label generators (HouseKeep: 119 eps.; ours: 201 eps.)
}
\label{tab:hk_labels}
\resizebox{\linewidth}{!}{%
\begin{tabular}{|l|c|c|c|c|}
\hline
 & \multicolumn{2}{c|}{\textbf{ECO/$N$}~$\downarrow$} & \multicolumn{2}{c|}{\textbf{TypeAcc}~$\uparrow$} \\
\cline{2-3}\cline{4-5}
\textbf{Method} & Ours & HouseKeep & Ours & HouseKeep \\
\hline
\textit{Constant: cabinets first} & \textit{.484} & \textit{\textbf{.248\,$\pm$\,.007}} & \textit{74.6\%} & \textit{\textbf{98.9\%}} \\
\hline
Random (5 seeds)   & .455 & .321\,$\pm$\,.015 & 63.3\% & 70.3\% \\
Visual CLIP        & .466 & .322\,$\pm$\,.025 & 60.2\% & 58.3\% \\
Text CLIP          & .436 & .301\,$\pm$\,.009 & 52.2\% & 43.4\% \\
\hline
TidyBot commonsense (zero-shot)  & .338 & .257\,$\pm$\,.012 & 92.0\% & 87.2\% \\
TidyBot summarization (informed) & .332 & .263\,$\pm$\,.008 & 89.6\% & 82.0\% \\
LLM scene graph (Haiku 4.5)      & .284 & .302\,$\pm$\,.006 & 88.6\% & 77.8\% \\
\hline
\textbf{LOCUS} ($K{=}2$, zero-shot on HK) & \textbf{.257} & .264\,$\pm$\,.004 & \textbf{95.2\%} & 82.8\% \\
\hline
\end{tabular}%
}
\end{table}

\subsubsection{Ablations}
We retrained all feature and loss ablations on iTHOR (Table~\ref{tab:mp_proximity}).
Removing the 4-d height-tier block from containers and landmarks ($-$height) raises ECO, suggesting that the height features contribute to the reduction in search cost seen in Table~\ref{tab:main}.
Message passing improves ECO by ${\sim}0.28$, with or without \texttt{soft-ldist}, as the graph carries landmark context, but results indicate the strongest performance comes from having both features. Uniform and $e^{-d}$ edge weights perform similarly, so finer geometry is unlikely to improve performance.

\begin{table}[h]
\centering
\small
\setlength{\tabcolsep}{3pt}
\caption{Message passing vs.\ proximity and height features, best checkpoint. Best per column in \textbf{bold}.}
\label{tab:mp_proximity}
\resizebox{\linewidth}{!}{%
\begin{tabular}{|l|c|c|c|c|}
\hline
\textbf{Condition} & \textbf{IRA}\,$\uparrow$ & \textbf{S-IRA}\,$\uparrow$ & \textbf{ECO}\,$\downarrow$ & \textbf{S-ECO}\,$\downarrow$ \\
\hline
$-$sldist, $K{=}0$ & .267 & .412 & 3.259 & 2.587 \\
$-$sldist, $K{=}2$, uniform & .333 & .442 & 2.706 & 2.259 \\
$-$sldist, $K{=}2$, $e^{-d}$ & .321 & .412 & 2.642 & 2.189 \\
\hline
$-$height, $K{=}0$ & .339 & .448 & 3.502 & 2.612 \\
$-$height, $K{=}2$ & .358 & .455 & 3.393 & 2.527 \\
\hline
Full, $K{=}0$ & \textbf{.406} & .491 & 2.776 & 2.259 \\
Full, $K{=}2$ (LOCUS) & .382 & \textbf{.497} & \textbf{2.478} & \textbf{2.119} \\
\hline
\end{tabular}}
\end{table}


\subsection{Sim-to-Sim Transfer: \mbox{RoboCasa}}

\subsubsection{Setup}

 Few simulators support placing objects inside closed containers, which limits dataset scale; to zero-shot test cross-domain generalization without real-world perception noise, we evaluate the iTHOR-trained model on RoboCasa~\cite{robocasa2024} kitchen scenes rendered in MuJoCo~\cite{todorov2012mujoco}. RoboCasa draws on a different object asset library and distribution of floor plans than iTHOR, yet accepts the same placement policy, so the environment shifts while the labeling convention is held fixed. Performance indicates that the learned spatial conventions transfer across simulators. We evaluate 10 layouts $\times$ 3 styles $\times$ 5 placement seeds, giving 1{,}188 episodes. RoboCasa scenes hold 20-40 openable containers--roughly 5 $\times$ iTHOR--making instance discrimination much harder.

\subsubsection{Results}

Table~\ref{tab:robocasa_main} evaluates LOCUS zero-shot: trained only on iTHOR, tested in RoboCasa kitchens with no finetuning. Despite the fivefold larger candidate set, LOCUS continues to beat the baselines, suggesting that the learned spatial priors transfer across kitchen simulators.



\begin{table}
\setlength{\tabcolsep}{3pt}
\small
\centering
\caption{Sim-to-sim transfer: iTHOR test set (201 episodes) against RoboCasa (10 layouts $\times$ 3 styles $\times$ 5 seeds, 1{,}188 episodes). LOCUS is trained on iTHOR only, with no finetuning.}
\label{tab:robocasa_main}
\begin{tabular}{|l|cc|cc|}
\hline
& \multicolumn{2}{c|}{\textbf{IRA}~$\uparrow$} & \multicolumn{2}{c|}{\textbf{ECO/$N$}~$\downarrow$} \\
\textbf{Method} & iTHOR & RoboCasa & iTHOR & RoboCasa \\
\hline
Random          & .241 & .037 & .455 & .507 \\
Visual CLIP     & .188 & .048 & .465 & .519 \\
Class Text CLIP & .182 & .075 & .436 & .373 \\
\hline
LOCUS ($K\!=\!2$) & \textbf{.367} & \textbf{.134\,$\pm$\,.031} & \textbf{.257} & \textbf{.314\,$\pm$\,.025} \\
\hline
\end{tabular}
\end{table}

\begin{figure*}[t!]
    \centering
    \resizebox{.8\linewidth}{!}{\begin{tikzpicture}[
    >=Stealth,
    node distance=0.6cm and 0.8cm,
    rosnode/.style={
        rectangle, rounded corners=4pt,
        draw=nodeblue, line width=1pt,
        fill=nodefill,
        minimum width=2.4cm, minimum height=0.75cm,
        font=\sffamily\bfseries\small,
        text=nodeblue,
        align=center
    },
    sgblock/.style={
        rectangle, rounded corners=4pt,
        draw=sggreen, line width=1pt,
        fill=sgfill,
        minimum width=2.2cm, minimum height=0.7cm,
        font=\sffamily\small,
        text=sggreen,
        align=center
    },
    sensor/.style={
        rectangle, rounded corners=2pt,
        draw=buildpurple, line width=0.6pt,
        fill=buildfill,
        minimum width=1.8cm, minimum height=0.6cm,
        font=\sffamily\footnotesize,
        text=buildpurple,
        align=center
    },
    innermodel/.style={
        rectangle, rounded corners=2pt,
        draw=sensorgray, line width=0.6pt,
        fill=sensorfill,
        minimum width=1.6cm, minimum height=0.55cm,
        font=\sffamily\footnotesize,
        text=sensorgray,
        align=center
    },
    querybox/.style={
        rectangle, rounded corners=4pt,
        draw=buildpurple, line width=1pt,
        fill=buildfill,
        minimum width=2.0cm, minimum height=0.7cm,
        font=\sffamily\bfseries\footnotesize,
        text=buildpurple,
        align=center
    },
    artifact/.style={
        diamond, aspect=2.0,
        draw=sggreen, line width=1.2pt,
        fill=sgfill,
        font=\sffamily\bfseries\small,
        text=sggreen,
        inner sep=2pt,
        align=center
    },
    data/.style={
        diamond, aspect=2.0,
        draw=topicamber, line width=1.2pt,
        fill=topicamber!8,
        minimum width=1.5cm, minimum height=0.75cm,
        font=\sffamily\bfseries\footnotesize,
        text=topicamber,
        inner sep=1.5pt,
        align=center
    },
    internal/.style={
        -{Stealth[length=5pt]},
        line width=0.8pt,
        sggreen,
    },
    feedin/.style={
        -{Stealth[length=5pt]},
        line width=0.6pt,
        buildpurple!70,
    },
    buildflow/.style={
        -{Stealth[length=6pt]},
        line width=1.2pt,
        buildpurple!70,
    },
    searchflow/.style={
        -{Stealth[length=6pt]},
        line width=1.2pt,
        buildpurple!70,
    },
    blabel/.style={
        font=\sffamily\scriptsize,
        text=buildpurple!70,
        inner sep=1.5pt, rounded corners=1pt,
    },
    slabel/.style={
        font=\sffamily\scriptsize,
        text=buildpurple!70,
        inner sep=1.5pt, rounded corners=1pt,
    },
]


\node[sgblock] (detic) {Detic Detector};
\node[sgblock, below=0.15cm of detic] (clip) {CLIP Encoder};
\node[sgblock, right=1.5cm of detic] (voxel) {Voxel Graph\\Builder};
\node[data, right=0.35cm of voxel] (sg) {Scene\\Graph};
\node[sgblock, right=0.35cm of sg] (gnn) {GNN Scorer};

\draw[internal] (detic.east) -- (voxel.west)
    node[font=\sffamily\scriptsize, text=sggreen, midway, above]{detections};
\draw[internal] (clip.east) -- (voxel.south west)
    node[font=\sffamily\scriptsize, text=sggreen, pos=0.6, below right]{embeddings};
\draw[internal] (voxel.east) -- (sg.west);
\draw[internal] (sg.east) -- (gnn.west);

\begin{pgfonlayer}{middle}
\node[fit=(detic)(clip)(voxel)(sg)(gnn),
      draw=nodeblue, line width=1pt,
      rounded corners=4pt, fill=nodefill,
      inner xsep=8pt, inner ysep=10pt,
      label={[font=\sffamily\bfseries\small, text=nodeblue, anchor=north west,
             inner sep=4pt]north west:Detection Node}] (detect) {};
\end{pgfonlayer}


\node[rosnode, above=0.5cm of detect.north, xshift=-1.0cm] (explore) {Exploration\\Node};

\draw[-{Stealth[length=6pt]}, line width=1.2pt, nodeblue!70] (explore.south) -- (explore.south |- detect.north)
    node[font=\sffamily\scriptsize, text=nodeblue!70, inner sep=1.5pt, midway, anchor=west, xshift=3pt]{at each viewpoint};


\node[data, right=0.35cm of detect.east] (rankings) {Container\\Rankings};

\node[rosnode, right=0.35cm of rankings] (navopen) {Navigate \&\\Open Node};


\node[sensor, above=0.35cm of detect.north west, xshift=1.6cm,
      minimum width=2.8cm, minimum height=1.1cm,
      align=left] (sensors) {%
    \textbf{Sensors}\\[1pt]
    {\footnotesize RGB Camera}\\
    {\footnotesize Depth Camera}\\
    {\footnotesize TF / Camera Pose}%
};

\draw[feedin] (sensors.south) -- (sensors.south |- detect.north);


\node[querybox, right=3.0cm of explore] (query) {Query\\{\normalfont\itshape e.g.\ ``fork''}};

\draw[searchflow] (query.south) -- (query.south |- detect.north)
    node[slabel, pos=0.5, anchor=west, xshift=3pt]{CLIP-encoded};


\draw[searchflow] (detect.east) -- (rankings.west);
\draw[searchflow] (rankings.east) -- (navopen.west);

\begin{scope}[on background layer]
    \node[fit=(explore)(detect)(sensors),
          draw=buildpurple!40, line width=0.8pt, dashed,
          rounded corners=8pt, fill=buildfill!15,
          inner sep=10pt,
          label={[font=\sffamily\bfseries\small, text=buildpurple]
                 above left:Scene Graph Building}] (buildbox) {};

    \node[fit=(query)(rankings)(navopen),
          draw=buildpurple!40, line width=0.8pt, dashed,
          rounded corners=8pt, fill=buildfill!15,
          inner sep=10pt,
          label={[font=\sffamily\bfseries\small, text=buildpurple]
                 above left:Search}] (searchbox) {};
\end{scope}

\node[anchor=north east, font=\sffamily\bfseries\small, text=sensorgray]
    at ([xshift=-0.9cm, yshift=3.2cm]navopen.north) (leg) {Legend};

\node[rectangle, rounded corners=2pt,
      draw=nodeblue!60, line width=0.5pt, fill=nodefill!50,
      minimum width=0.4cm, minimum height=0.22cm,
      anchor=north west]
    at ([xshift=0.1cm, yshift=-0.15cm]leg.south west) (leg_bnode) {};
\node[font=\sffamily\footnotesize, text=sensorgray!70, anchor=west]
    at ([xshift=0.08cm]leg_bnode.east) (leg_bnode_text) {ROS\,2 Node};

\node[rectangle, rounded corners=2pt,
      draw=sggreen!60, line width=0.5pt, fill=sgfill!50,
      minimum width=0.4cm, minimum height=0.22cm,
      anchor=north west]
    at ([yshift=-0.1cm]leg_bnode.south west) (leg_proc) {};
\node[font=\sffamily\footnotesize, text=sensorgray!70, anchor=west]
    at ([xshift=0.08cm]leg_proc.east) (leg_proc_text) {Data Processors};

\node[rectangle, rounded corners=1pt,
      draw=buildpurple!60, line width=0.5pt, fill=buildfill!50,
      minimum width=0.4cm, minimum height=0.22cm,
      anchor=north west]
    at ([yshift=-0.1cm]leg_proc.south west) (leg_sensor) {};
\node[font=\sffamily\footnotesize, text=sensorgray!70, anchor=west]
    at ([xshift=0.08cm]leg_sensor.east) (leg_sensor_text) {Input};

\node[rectangle, rounded corners=2pt,
      draw=buildpurple!60, line width=0.5pt, dashed, fill=buildfill!30,
      minimum width=0.4cm, minimum height=0.22cm,
      anchor=north west]
    at ([yshift=-0.1cm]leg_sensor.south west) (leg_method) {};
\node[font=\sffamily\footnotesize, text=sensorgray!70, anchor=west]
    at ([xshift=0.08cm]leg_method.east) (leg_method_text) {Pipeline Steps};

\node[diamond, aspect=1.8,
    draw=topicamber!60, line width=0.5pt, fill=topicamber!8,
    inner sep=2pt,
    anchor=north west]
  at ([xshift=2.5pt, yshift=-0.1cm]leg_method.south west) (leg_data) {};
\node[font=\sffamily\footnotesize, text=sensorgray!70, anchor=west]
  at ([xshift=0.08cm]leg_data.east) (leg_data_text) {Data Artifact};

\begin{pgfonlayer}{background}
\node[fit=(leg)(leg_bnode)(leg_bnode_text)(leg_proc)(leg_proc_text)(leg_sensor)(leg_sensor_text)(leg_method)(leg_method_text)(leg_data)(leg_data_text),
      rectangle, rounded corners=3pt,
      draw=sensorgray, line width=0.6pt,
      fill=sensorfill,
      inner sep=5pt] {};
\end{pgfonlayer}

\end{tikzpicture}}
    \caption{ROS pipeline. The scene graph is built during exploration then fed to the GNN.}
    \label{fig:ros_pipeline}
\end{figure*}
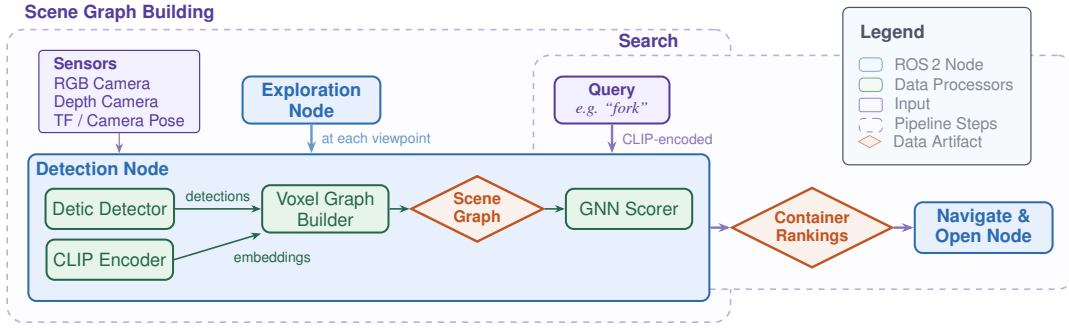

\begin{table}
\setlength{\tabcolsep}{3pt}
\small
\centering
\caption{In-distribution vs.\ novel (OOD) query split in RoboCasa.}
\label{tab:robocasa_dist}
\begin{tabular}{|l|l|c|c|}
\hline
\textbf{Method} & \textbf{Split} & \textbf{IRA}~$\uparrow$ & \textbf{ECO/$N$}~$\downarrow$ \\
\hline
Random         & In-dist & 0.023 & 0.515 \\
               & Novel   & 0.032 & 0.542 \\
\hline
Visual CLIP    & In-dist & 0.049 & 0.492 \\
               & Novel   & 0.035 & 0.545 \\
\hline
Class Text CLIP & In-dist & 0.147 & 0.273 \\
               & Novel   & 0.000 & 0.485 \\
\hline
LOCUS,  & In-dist & 0.139 & 0.306 \\
               & Novel   &  0.129 & 0.323 \\
\hline
\end{tabular}
\end{table}

Table~\ref{tab:robocasa_dist} splits results by query novelty and shows LOCUS generalizes to unseen object categories without finetuning. LOCUS maintains consistent performance across both splits, demonstrating that the graph-based spatial reasoning LOCUS learns generalizes beyond the training vocabulary. 
Additional in distribution (ID) and out-of-distribution (OOD) analysis per-query determined LOCUS achieves the best IRA on 14 of 17 Robocasa queries. Strong performance on utensils (fork, tongs, ladle) and food (shrimp, mustard) confirms that spatial conventions (e.g., utensils near countertops, perishables in the fridge) transfer across kitchen simulators.


\subsection{Mobile Manipulator Hardware}
\label{mobilemanipulatorresults}

\subsubsection{Setup}


 To show that our GNN, trained in simulation, works with a domain shift, we implemented our approach on a Hello Robot Stretch~3 mobile manipulator and tested it in a real kitchen. We used a deliberately cluttered environment, thereby adding extraneous detections as additional scene-graph nodes to test real-world robustness. We use Detic~\cite{zhou2022detic} for open-vocabulary object detection and extrapolate 3D poses from bounding boxes and the associated depth image. We deduplicate overlapping detections by confidence, keeping the highest-scoring box, and use the surviving poses and labels to generate nodes. Nodes whose labels match a container or handle class in our ontology are labeled openable; all other nodes serve as spatial context. The GNN ingests and ranks the full scene graph, and the robot searches the ranked openable nodes. Fig.~\ref{fig:ros_pipeline} shows the end-to-end pipeline.

\begin{figure}
 \centering
 \begin{subfigure}{0.49\linewidth}
 \centering
  \includegraphics[width=.9725\textwidth]{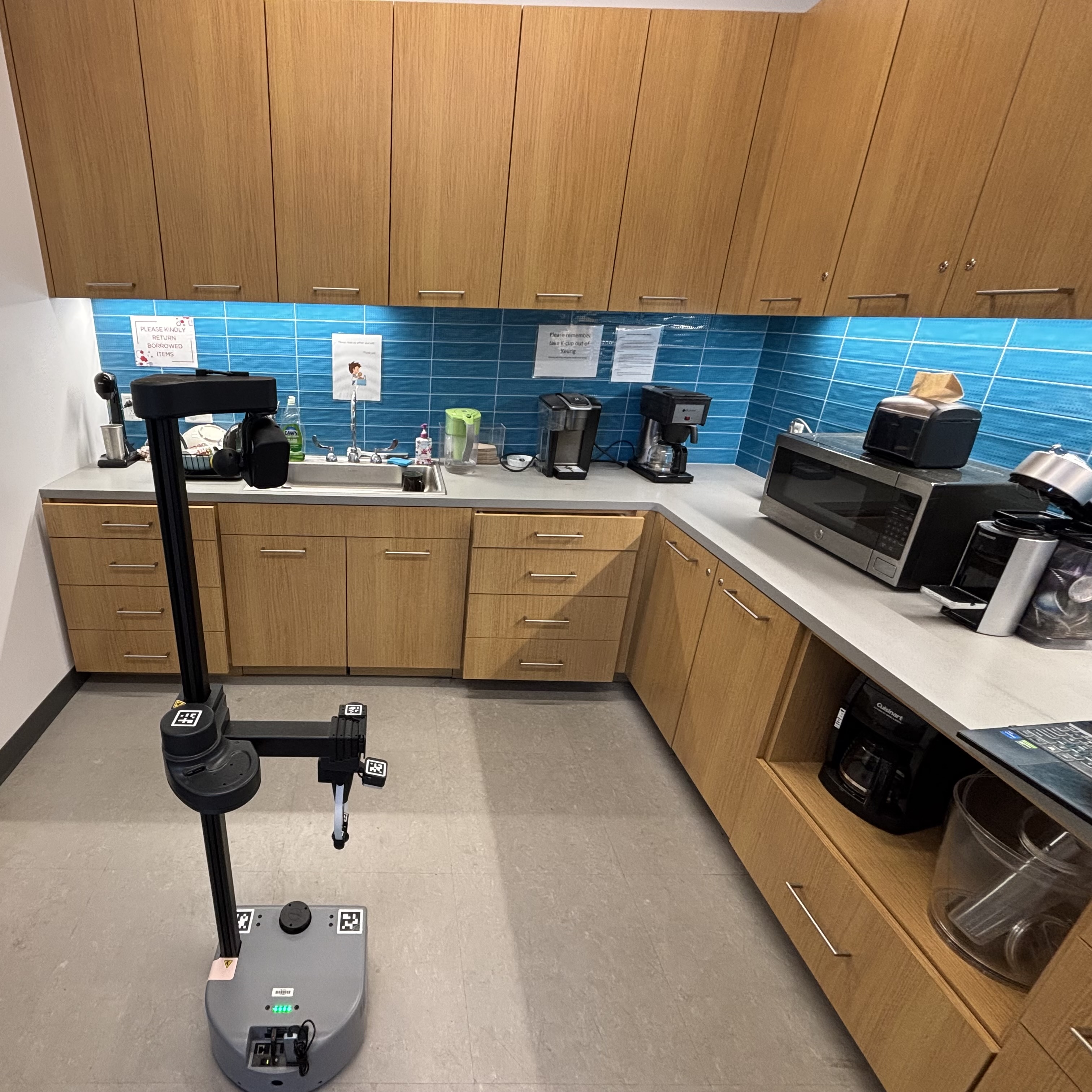}
 \caption{Test kitchen and Stretch~3.}
 \label{fig:ros_kitchen}
 \end{subfigure}\hfill
 \begin{subfigure}{0.49\linewidth}
 \centering
 \includegraphics[width=\textwidth]{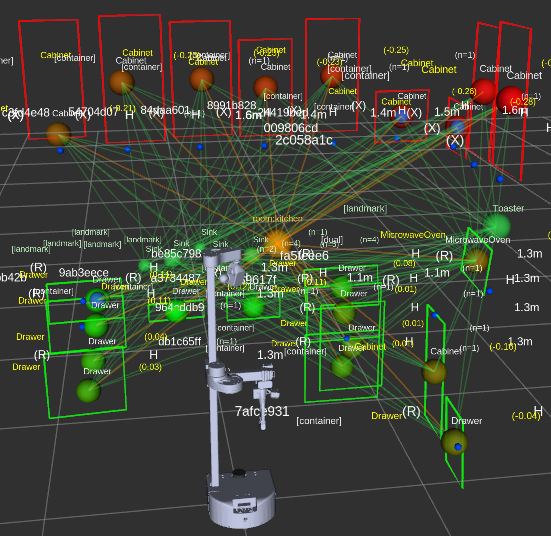}
 \caption{Scene graph.}
 \label{fig:ros_sg_only}
 \end{subfigure}
 \caption{The physical kitchen used for the hardware trials, and the scene graph built by Stretch-3 after exploration. Ranked scene graph nodes (green:best) and node reachability (green:reachable, red:unreachable).}
\label{fig:ros_sg}
\end{figure}

We average 3 runs per query–method combination from a fixed start, panning the camera to map the kitchen, for the queries fork, mug, and sponge. Fig.~\ref{fig:ros_sg} shows the test kitchen and a completed scene graph.

Building the scene graph from real detections introduces perception noise absent from simulation (See Fig \ref{fig:ros_sg}): estimated rather than ground-truth poses, sensor noise in depth readings, and reduced frame rates from ROS message throttling over rosbridge, which limits the number of frames available to the labeler and can degrade detection quality at oblique viewing angles.

\subsubsection{Results}

\begin{table}[h]
\setlength{\tabcolsep}{3pt}
\small
\centering
\caption{Per-query Stretch~3 in a real kitchen.}
\label{stretch_results:tab}
\begin{tabular}{|l|c|c|c|}
\hline
& \multicolumn{3}{c|}{\textbf{ECO} $\downarrow$} \\
\cline{2-4}
\textbf{Query} & Text CLIP & Vis.\ CLIP & LOCUS \\
\hline
Fork   & 11.0 & \textbf{1.0} & \textbf{1.0} \\
\hline
Mug    & 2.0  & 3.0          & \textbf{1.0} \\
\hline
Sponge & 8.5  & \textbf{4.0} & 6.5 \\
\hline
\end{tabular}
\end{table}

Table ~\ref{stretch_results:tab} shows that on average LOCUS performs competitively. For the query sponge, Detic mislabels the cabinet that contains the sponge as a drawer. This impacts both LOCUS and Text CLIP, but our spatial reasoning yields ``smarter mistakes'' under real-world mislabeling.


\section{Conclusion}
\label{sec:conclusion}
To our knowledge, LOCUS is the first approach enabling a robot to choose among containers, including same-class instances, to retrieve an occluded object. Our results show our GNN updated CLIP embeddings transfer across simulators and to hardware. \textbf{Limitations} LOCUS ranks a fully built scene graph rather than a partial one, so it requires an exploration pre-step. Queries with weak spatial signal (e.g., pots and pans, whose placement varies across kitchens) stay near-random, and landmark-sparse rooms like living rooms offer little signal for message passing. 
 No public dataset provides instance-level object-in-container ground truth across diverse room classes; prior rearrangement work~\cite{kant2022housekeep} gives only class-level receptacle preferences. We therefore built our own from AI2-THOR using  a hand-crafted placement policy that encodes commonsense conventions but not the full variability of real homes. 



\bibliographystyle{ieeetr}
\bibliography{main}  

\end{document}